\documentclass{article}
\usepackage{graphicx}
\usepackage{latexsym}
\usepackage{amsmath,amssymb}
\usepackage{graphicx}
\usepackage{times}
\usepackage[utf8]{inputenc}
\usepackage{verbatim}
\usepackage{multicol}
\usepackage{url}

\usepackage{caption}
\usepackage{tkz-tab}
\usepackage{psfrag}

\usepackage{pgfplots}
\usepackage{textcomp}
\usetikzlibrary{matrix}
\usetikzlibrary{positioning,shapes,shadows,arrows,backgrounds}
\usepackage{color}

\usepackage[abs]{overpic}
\usepackage{pict2e}

\newcommand{\B}{\mathcal{B}} 
\newcommand{\T}{^\mathsf{T}}
\newcommand{\mM}{{\mathbf{M}}}
\newcommand{\vgamma}{{\boldsymbol{\gamma}}}
\newcommand{\vlambda}{{\boldsymbol{\lambda}}}
\newcommand{\vq}{{\mathbf{q}}}
\newcommand{\vdq}{{\mathbf{\dot{q}}}}
\newcommand{\vddq}{{\mathbf{\ddot{q}}}}
\newcommand{\vs}{{\mathbf{s}}}
\newcommand{\vds}{{\mathbf{\dot{s}}}}
\newcommand{\vdds}{{\mathbf{\ddot{s}}}}
\newcommand{\vzero}{\mathbf{0}}
\newcommand{\mdPhidq}{\dot{\boldsymbol{\phi}}_{\vdq}}
\newcommand{\mdPhidqT}{\dot{\boldsymbol{\phi}}_{\vdq}{\T}}

\newcommand{\vphin}{{\boldsymbol{\phi}^n}}
\newcommand{\vphid}{{\boldsymbol{\phi}^d}}
\newcommand{\vdelta}{{\boldsymbol{\delta}}}
\newcommand{\vbeta}{{\boldsymbol{\beta}}}
\newcommand{\vgamman}{{\boldsymbol{\gamma}^n}}

\newcommand{\mdPhiddq}{\dot{\boldsymbol{\phi}}_{\vdq}^{d}}
\newcommand{\mdPhindq}{\dot{\boldsymbol{\phi}}_{\vdq}^{n}}
\newcommand{\mdPhidds}{\dot{\boldsymbol{\phi}}_{\vds}^{d}}
\newcommand{\mdPhinds}{\dot{\boldsymbol{\phi}}_{\vds}^{n}}
\newcommand{\mdPhiddqT}{\dot{\boldsymbol{\phi}}_{\vdq}^{d}{\T}}
\newcommand{\mdPhindqT}{\dot{\boldsymbol{\phi}}_{\vdq}^{n}{\T}}
\newcommand{\mdPhiddsT}{\dot{\boldsymbol{\phi}}_{\vds}^{d}{\T}}

\newcommand{\matriz}[1]{\mathbf{#1}}
\newcommand{\mR}[2]{{\matriz{R}_{#1}^{#2}}}
\newcommand{\vect}[1]{\boldsymbol{#1}}
\begin{document}

\title{Symbolic Multibody Methods for Real-Time Simulation of Railway Vehicles.}



\author{
        Javier Ros   \and
        Aitor Plaza  \and
        Xabier Iriarte \and
        Jesús María Pintor
}


\date{Department of Mechanical Engineering\\ Public University of Navarre \\ Campus de Arrosadia, 31006 Pamplona, Navarre, Spain \\ \textit{[jros,aitor.plaza,xabier.iriarte,txma]@unavarra.es}}

\maketitle

\begin{abstract}
In this work,  recently developed state-of-the-art \textit{symbolic multibody} methods are tested to acurately model a complex railway vehicle. The model is generated using a symbolic implementation of the principle of the virtual power. Creep forces are  modeled using a direct symbolic implementation of the standard linear Kalker model. No simplifications, as base parameter reduction, partial-linearization or look-up tables for contact kinematics, are used. An  Implicit-Explicit integration scheme is proposed to efficiently deal with the stiff creep dynamics. Hard real-time performance is achieved: the CPU time required for a very stable $1 ~\text{ms}$ integration time step is $256 ~\mu\text{s}$.

\end{abstract}

\section{Introduction}
\label{intro}
Multibody system dynamics is a well established discipline in the context of railway vehicle design.
 It is used for new concept performance evaluation, stability, lifetime, wear prediction, etc. In general it is desirable to be able to do these analyses as fast as possible. In particular, due to the huge number of computations required, computational performance can be very important when dealing with design optimization. Nevertheless these tasks do not demand strict real-time performance.
 
 The computational power available on today's off-the-shelf computers is getting closer to allow real-time direct numerical simulation of complex railway vehicle models. This in turn opens up new possibilities
 that can greatly benefit the design, safety, and model based predictive maintenance in the railway field.
 Most important applications can be considered to be HiL (Hardware in the Loop) on the design side, and on-line filtering techniques (Kalman filter alike) in the context of safety, and model based predictive maintenance.
  These developments usually run on the heels of previous work done in the context of vehicle dynamics.
 
 Symbolic multibody models have demonstrated to be an effective tool for the modeling of general multibody systems. In particular they have been shown to be very fast when using recursive $O(n^3)$ formulations \cite{samin2003}, that in turn require a parametrization based on relative coordinates. Main challenges are related to the enormous size of the expressions that the symbolic processor needs to deal with as this can limit the size of the problem to be analyzed. Recently, in \cite{Plaza2015} the authors presented a symbolic multibody library in which the concept of recursivity is extended so that it is no longer based on the formulation but, instead, on the parametrization level. This is achieved by the definition of an algebra that includes the typical mechanics operators (position vector, velocity,... ) and that deals with the recursivity that might be embedded into the parametrization. The typical tree-shaped body structure is replaced by a tree structure for points and another one for bases. This gives a fine grained control of the recursivity that, in this way, can be different for both tree structures.  No limitation is imposed on the parametrization of the system. As a consequence the library allows to implement arbitrary dynamics formulations. Atomization (optimization of symbolic expression representation) is embedded into the library from the very bottom upwards. This  alleviates the symbolic manipulation of expressions and  lowers their complexity to a minimum. This in turn allows to obtain optimal atomizations that minimize the computational complexity and increases the size of the problems that is possible to analyze.
 
This article pursues to evaluate the feasibility of real-time numerical simulation of a complex locomotive multibody model  using  state of the art symbolic modeling techniques referred above. For our study we use the FEVE 3000 locomotive. A generic (spline based) definition for the contact surfaces of the wheels and rails, including irregularities, is used. Based on these, creep forces are modeled using a direct symbolic implementation of the standard linear Kalker model without simplifications of any kind. Bodies and rail are considered rigid with three-dimensional kinematics. No further simplifications as contact coordinate removal \cite{Shabana2005}, pre-calculated tables \cite{malvezzi2008determination}, partial linearization \cite{Escalona2015} or base parameter reduction \cite{Iriarte2015} are presented.

To that end the modeling is done based on the multibody system symbolic library \verb!lib_3D_MEC_GiNaC! \cite{Plaza2015}, using a relative  parametrization with respect to the inertial reference. 

The paper is structured  as follows: In section \ref{sec1} the symbolic methods used in this work are briefly described. In section \ref{sec2} the description of the modeled system is presented. In section \ref{sec3} the most interesting details of the multibody modeling are presented. In section \ref{sec4} the results of the simulations are shown and discussed. Finally in section \ref{sec5} the main conclusions of this work are presented.

\section{Symbolic modeling procedures}
\label{sec1}
Simply stated, the main goal of the symbolic modeling of multibody systems can be defined as: 

``\textit{to obtain a set of functions that allow for the determination of the position, velocities and accelerations of all the bodies of the system}''.

Special symbolic procedures are required if a real-time-capable fast multibody model is desired. The main features of these procedures, as
proposed in \cite{PlazaThesis2015}, are summarized below.

\subsection{Parametrization and system topology.}

\begin{figure}
\centering
\begin{minipage}{.33\textwidth}
  \centering
  \includegraphics[width=\textwidth]{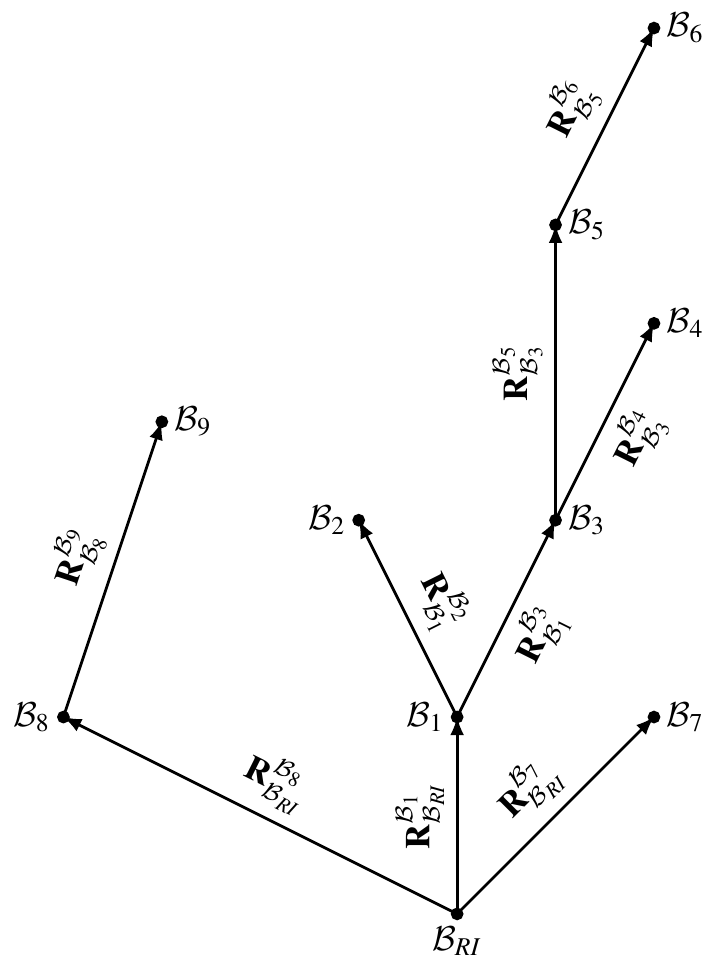}
\end{minipage}%
\begin{minipage}{.05\textwidth}
\phantom{A}
\end{minipage}%
\begin{minipage}{.6\textwidth}
  \centering
  \includegraphics[width=\textwidth]{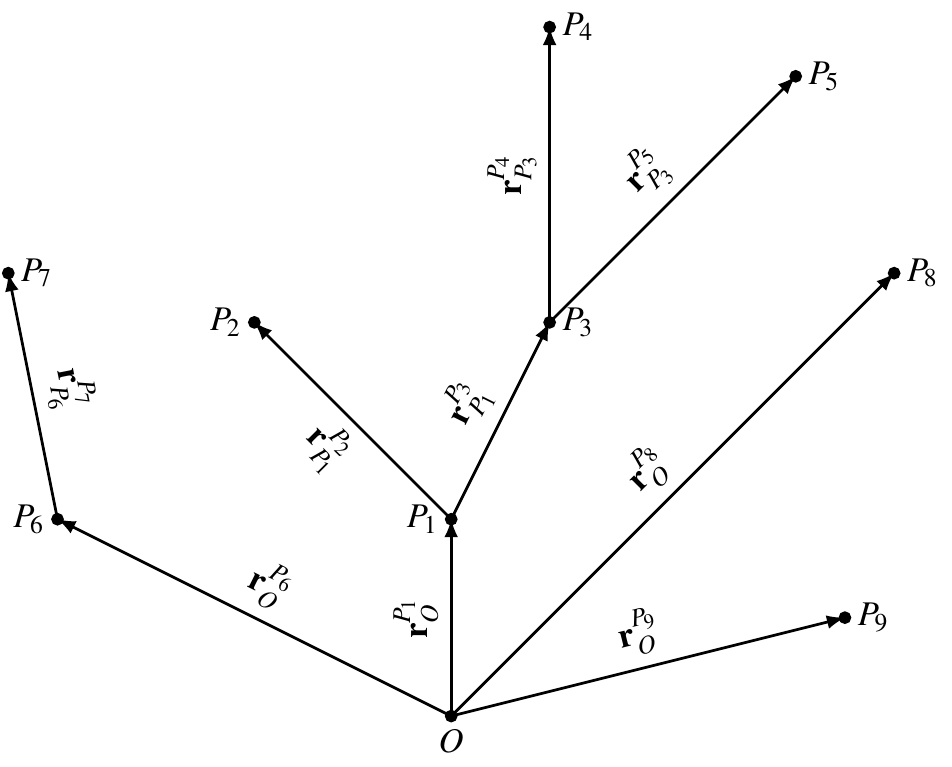}
\end{minipage}
\caption{Illustrative examples of bases (left) and points (right) structures }
\label{fig:points_bases}
\end{figure}

In order to model the multibody system a set of geometric parameters $\mathbf{p}$ and generalized coordinates $\mathbf{q}$, along with their associated velocity $\dot{\mathbf{q}}$ and generalized accelerations $\ddot{\mathbf{q}}$ are defined. We propose to split up the classical tree-shaped body structure into two different tree-shaped structures: 1) the bases structure and 2) the points structure, see Fig.~\ref{fig:points_bases}.

In this approach, bases $\B_{j}$ and points $P_{j}$ are defined in terms of other bases $\B_{i}$ and points $P_{i}$ by the way of relative base-change or rotation matrices  $\mathbf{R}_{\B_{i}}^{\B_{j}}$ and positions vectors $\mathbf{r}_{P_{i}}^{P_{j}}$. The functions used by the symbolic library \cite{ros2007lib3d} can be schematically represented as
\begin{eqnarray}
\label{eq:new_base}
\B_{i} &\xrightarrow{\mathbf{R}_{\B_{i}}^{\B_{j}}(exp_x(\mathbf{q},~t,\mathbf{p}),exp_y(\mathbf{q},t,\mathbf{p}),exp_z(\mathbf{q},t,\mathbf{p}),~exp_\phi(\mathbf{q},t,\mathbf{p}))} &\B_{j}~~~~~~~ \\
\label{eq:new_point}
P_{i} &\xrightarrow{\mathbf{r}_{P_{i}}^{P_{j}}(exp_x(\mathbf{q},t,\mathbf{p}),~exp_y(\mathbf{q},t,\mathbf{p}),~exp_z(\mathbf{q},t,\mathbf{p}),~\B_k)}& P_{j},
\end{eqnarray}
%
%
where $exp_*(\mathbf{q},t,\mathbf{p})$ represent arbitrary symbolic expressions in terms of which vectors and base-change matrices are defined. This in turn confers physical meaning to the defined coordinates and parameters. Note that, in order to illustrate the procedure, the rotation matrix appearing in Eq.~(\ref{eq:new_base}) is parametrized using Euler parameters.

This splitting of the body structure into the bases and points structures confers complete flexibility to the choice of the parametrization. A body position and orientation
no longer needs to be defined with respect to the preceding body in the tree-shaped bodies structure.
Instead, the body position is given by a point in the points structure and an orientation by a base in the bases structure. 

It should be noted that the bases structure is independent of the points structure. Conversely, the points structure is dependent on the bases structure as the relative position vector components are given using arbitrary bases (note the $\B_k$ parameter in Eq.~(\ref{eq:new_point})). Finally, it should be understood that the nature of the parameterization depends on the definition of the rotation matrices and position vectors. For example, if they are defined with respect to another point or orientation the coordinates will be ``relative'', but if they are defined with respect to an absolute point or orientation they will be ``absolute''.

\subsection{``On-the-way'' atomization}

The symbolic expressions that we need to deal with can be huge. The successive multiplication of symbolic expressions leads to an explosive growth in the size of the expressions that can limit the maximum size of the multibody systems that can be analyzed. In order to deal with this problem we use a standard technique in the context of symbolic computations that we call \textit{atomization}.

Atomization is a technique that condenses a symbolic expression set by splitting their expressions into several elemental sub-expressions. We call ``atoms'' to these elemental sub-expressions. They can be defined in terms of binary operations between symbols, numbers and/or other atoms. Or as transcendental functions of atoms.

This technique is beneficial when repeated sub-expressions appear and the same atom is used to represent them. Symbolically, this means less memory -as the sub-expression is allocated in memory once- and faster
symbolic manipulations. Numerically, this implies that the repeated sub-expression is only computed once leading to computational cost savings.

In the present context, we deal with the sets of expressions related to
the functions used to computationally implement a given MSD formalism. This atomized representation leads directly to the exportation of these functions in a way that benefits from the referred computational cost savings. To get a less abstract idea, Fig.~\ref{fig_atomization} shows the exported C code for the mass matrix of a simple four-bar linkage mechanism.

\begin{figure}
\hrule
\begin{multicols}{2}
{\tiny
\begin{verbatim}
atom27 = sin(theta2);
atom0 = cos(theta3);
atom1 = sin(theta3);
atom26 = cos(theta2);
atom49 =  atom1*atom26+atom27*atom0;
atom46 = -atom27*atom1+atom26*atom0;
atom237 = m3*l2*( cg3x*atom0+atom1*cg3z);
atom253 =  m3*l1*( cg3x*atom46+atom49*cg3z)+atom237;
atom200 = -l1*atom26;
atom214 = -l2*atom200;
atom197 = (l1*l1);
atom215 = (l2*l2);
atom270 =  atom27*l1*cg2z*m2+m3*( atom214+atom215)
           -m2*cg2x*atom200+I3yy+atom237+I2yy+atom253;
atom229 =  m3*l2*cg3x*atom0+m3*l2*atom1*cg3z;
atom271 =  m3*l1*cg3x*atom46+I3yy+m3*atom49*l1*cg3z
           +atom229;
atom273 =  I3yy+atom229;

_M[0] = -m3*( atom197+2.0*atom214+atom215)-atom197*m2
        -I3yy-I2yy+-2.0*atom253-I1yy
        +-2.0*( atom27*cg2z+atom26*cg2x)*l1*m2;
_M[1] = -atom270;
_M[2] = -atom271;
_M[3] = -atom270;
_M[4] = -m3*atom215-I3yy+-2.0*atom237-I2yy;
_M[5] = -atom273;
_M[6] = -atom271;
_M[7] = -atom273;
_M[8] = -I3yy;
\end{verbatim}
}
\end{multicols}
\hrule
\caption{Exported C code for atomized mass matrix.}
\label{fig_atomization}
\end{figure}

The atomization process should ideally be done ``on-the-way'', meaning that every time a new algebraic operation is performed a new atom is created or replaced by an existing  matching atom. 
Thus, the symbolic method takes advantage of the memory savings and the associated complexity reduction as soon as possible in the problem setup.
This means that the symbolic algebra system works internally with atomized expressions. A feature that is not obvious for the standard user but that is widespread in computer algebra systems. See Fig.~\ref{fig:on_the_way} for an elemental example.

\begin{figure}

\hrule
The addition of vectors $\vect{u}$ and $\vect{v}$ given their components represented in bases $\B_1$ and $\B_2$ 
\begin{equation*}
\begin{aligned}
\begin{Bmatrix} \vect{u} \end{Bmatrix}_{\B_1}= \begin{Bmatrix} u_x\\ u_y \\ u_z \end{Bmatrix}_{\B_1}
~~\text{and }~~
\begin{Bmatrix} \vect{v} \end{Bmatrix}_{\B_2}= \begin{Bmatrix} v_x\\ v_y \\ v_z \end{Bmatrix}_{\B_2}
\end{aligned}
\end{equation*}
is sought. Let the rotation matrix be
\begin{equation*}
\begin{aligned}
\mR{\B_1}{\B_2} = 
\begin{bmatrix}
1 &  0&0\\ 
0 &\cos(\theta)  & -\sin(\theta)\\ 
0 & \sin(\theta) & \cos(\theta)
\end{bmatrix}
\end{aligned}.
\end{equation*}
The addition of the two vectors represented in $\B_1$ base is performed as follows:
\begin{equation*}
\begin{aligned}
\begin{Bmatrix}\vect{u}+ \vect{u}\end{Bmatrix}_{\B_1} &=\begin{Bmatrix} \vect{u} \end{Bmatrix}_{\B_1} +\mR{\B_1}{\B_2} \begin{Bmatrix} \vect{v} \end{Bmatrix}_{\B_2} =
\begin{Bmatrix} u_x\\ u_y \\ u_z \end{Bmatrix}_{\B_1} +
\begin{bmatrix} 1 &  0&0\\0 &\cos(\theta)  & -\sin(\theta)\\ 0 & \sin(\theta) & \cos(\theta)\end{bmatrix}
\begin{Bmatrix} v_x\\ v_y \\ v_z \end{Bmatrix}_{\B_2}
=\\
&\begin{Bmatrix} u_x\\ u_y \\ u_z \end{Bmatrix}_{\B_1} +
\begin{bmatrix} 1 &  0&0\\0 &\alpha_1  & -\alpha_2 \\ 0 & \alpha_2  & \alpha_1 \end{bmatrix}
\begin{Bmatrix} v_x\\ v_y \\ v_z \end{Bmatrix}_{\B_2}
= 
\begin{Bmatrix} u_x\\ u_y \\ u_z \end{Bmatrix}_{\B_1} +
\begin{Bmatrix} v_x\\ \alpha_1 v_y -\alpha_2 v_z  \\ \alpha_2 v_y +\alpha_1 v_z \end{Bmatrix}_{\B_1}
=\\
&\begin{Bmatrix} u_x\\ u_y \\ u_z \end{Bmatrix}_{\B_1} +
\begin{Bmatrix} v_x\\ \alpha_3 - \alpha_4 \\ \alpha_5 + \alpha_6 \end{Bmatrix}_{\B_1}
=
\begin{Bmatrix} u_x\\ u_y \\ u_z \end{Bmatrix}_{\B_1} +
\begin{Bmatrix} v_x\\ \alpha_7 \\ \alpha_8 \end{Bmatrix}_{\B_1}
=
\begin{Bmatrix} \alpha_9\\ \alpha_{10} \\\alpha_{11} \end{Bmatrix}_{\B_1}
\end{aligned}
\end{equation*}
where $\alpha_i$ are the atoms:
\begin{equation*}
\begin{aligned}
\alpha_1  &= \cos(\theta) & \alpha_2  &= \sin(\theta)        & \alpha_3  &=\alpha_1 v_y         & \alpha_4 &= \alpha_2 v_z & \alpha_5  &=\alpha_2 v_y  &\alpha_6  =\alpha_1 v_z   \\
\alpha_7 &=\alpha_3 - \alpha_4  & \alpha_8  &= \alpha_5 - \alpha_6 & \alpha_9 &= u_x+v_x & \alpha_{10}  &=u_y  + \alpha_7  & \alpha_{11}  &= u_z + \alpha_8\\
\end{aligned}
\end{equation*}
\hrule
\caption{``On-the-way'' atomization example.}
\label{fig:on_the_way}
\end{figure}
In the same line, it is important to remember that the fundamental symbolic differentiation and substitution operations should be implemented to work directly on atomized expressions. This maximizes atom recycling and limits enormously the time and memory requirements of the algorithms.

In this context, to take advantage of the atomization, care should be taken when choosing the way and order in which the required operations
are performed.
The operation number should be minimized and atom recycling maximized.
A general purpose algorithm
aiming at finding an absolute minimum number of operations would require an exhaustive search that is beyond the reach of reasonable computational resources.
Therefore, it is required to define appropriate heuristics. Recursive dynamics formulations are usually taken as the starting point to define such heuristics. In our work, these heuristics are partly implemented by the way of mechanics operators, as will be explained in the next section.



\subsection{Recursive kinematic operators}

Recursive  formulations represent the state-of-the-art on symbolic MSD \cite{samin2003}. These formulations use relative coordinates to parametrize the system leading to a tree-shaped body structure\footnote{Closed loops are opened to parametrize and closed through constraint equations.}. 
This allows the recursive determination positions, velocities and accelerations of points as well as orientations, angular velocities and accelerations of bodies, by the way of the well known ``motion composition laws''. When different elements -points and orientations- share a common path towards the tree root, this implies the sharing of common sub-expressions. If applied symbolically, this recursive computation produces nearly good optimal ``on-the-way'' atomizations. This sharing of expressions is the main feature on which the  kinematic forward recursion step, found in recursive formulations, is based.

For example, for a serial multibody system the angular velocity of body $S_{i+1}$  with respect to $S_{i-1}$  could be expressed as follows:
\begin{eqnarray}
 \boldsymbol{\omega}^{S_{i+1}}_{S_{i-1}} = \boldsymbol{\omega}^{S_{i}}_{S_{i-1}} + \boldsymbol{\omega}^{S_{i+1}}_{S_{i}} 
\end{eqnarray}
In the same way the angular velocity of  body $S_{i+2}$ with respect to $S_{i-1}$ is expressed as:
\begin{eqnarray}
 \boldsymbol{\omega}^{S_{i+2}}_{S_{i-1}} = 
  \boldsymbol{\omega}^{S_{i}}_{S_{i-1}} + \boldsymbol{\omega}^{S_{i+1}}_{S_{i}} + \boldsymbol{\omega}^{S_{i+2}}_{S_{i+1}}=
 \boldsymbol{\omega}^{S_{i+1}}_{S_{i-1}} + \boldsymbol{\omega}^{S_{i+2}}_{S_{i+1}}
\end{eqnarray}
So, when computing magnitudes related to a given element it can be appreciated how computations related to elements down in the same chain can be reused. Other kinematic entities like position vectors, base-change matrices, linear velocities, accelerations and angular accelerations can be dealt with analogously. 

In correspondence with the substitution of the bodies structure by the bases and points structures proposed in our work, the recursivity at the level of bodies is now dealt with at the  bases and points structure levels. 
This allows not only to use arbitrary parametrizations, as commented before, but also a better use of the any degree of recursivity that may be implicit when using arbitrary parametrizations. 

To that end, kinematic operators that take advantage of any recursivity  present in the parameterization are defined: Position vector between two points, velocity of a point with respect to a given frame (point plus orientation), angular velocity, base-change matrix, and so on. Basically the typical \textit{recursivity} found in recursive algorithms is translated to the operator algebra. 

To support this an algebra of 3D vectors and tensors is defined. This algebra relieves the user of dealing with base-changes that are internally dealt with.
The full system works using ``on-the-way'' atomization and the operators are implemented taking advantage of the aforementioned recursivity. In this way, the number of operations is minimized and the reuse of atoms is maximized. As a consequence an optimal implementation of the given formalism for any  parameterization chosen by the user is obtained.



The backward recursion of $O(n^3)$ algorithms can be considered a particular implementation of the principle of the virtual power.
 The inertia forces and moments of the bodies affected by a given virtual movement appear added together in the contribution of this virtual movement to the system dynamic equations. Recursive formulations take advantage of this grouping so that they minimize the required operation count.
Taking advantage of this when applying symbolically  the principle of virtual power produces atomizations as efficient as state-of-the-art  $O(n^3)$ formulations. This is the approach followed by the symbolic implementation of the virtual power principle used in this work.

As an illustration of the achievements of our symbolic methods, we can obtain nearly optimal atomized equations for standard multibody systems using for example the principle of the virtual power and relative coordinates. Some authors \cite{samin2003} claim to be unable to do the same unless  a direct symbolic implementation of a recursive formulation is used\footnote{In comparison, our method presents the overhead of having to deal with the common atom search. Even if using hash tables to do the search our symbolic processing phase seems to take longer.}.

\subsection{Other symbolic methods}
There are other symbolic methods that can be applied to reduce even further the complexity of the resulting model: \textit{``trigonometricaly simplifiable expression removal''} \cite{PlazaThesis2015}, \textit{``base parameter formulation of the system inertias''}\cite{khalil1987,Ros2012,Ros2015}, \textit{``base parameter elimination''}\cite{Iriarte2015}, etc... This methods can be  applied directly on top of the presented modeling techniques. However, they are not considered in this work.

\section{Multibody model description}
\label{sec2}

\begin{figure}[h]
\centering
\includegraphics[width=\textwidth]{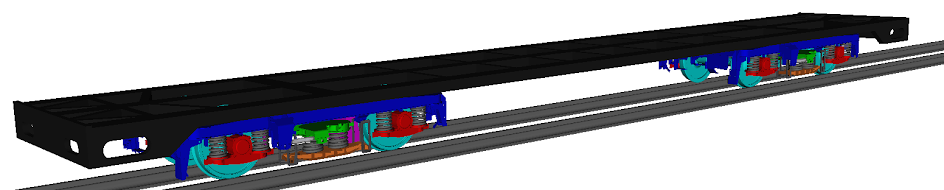}
\caption{Multibody model}
\label{fig:locomotive}
\end{figure}

\begin{figure}
\centering
 \includegraphics[width=\textwidth]{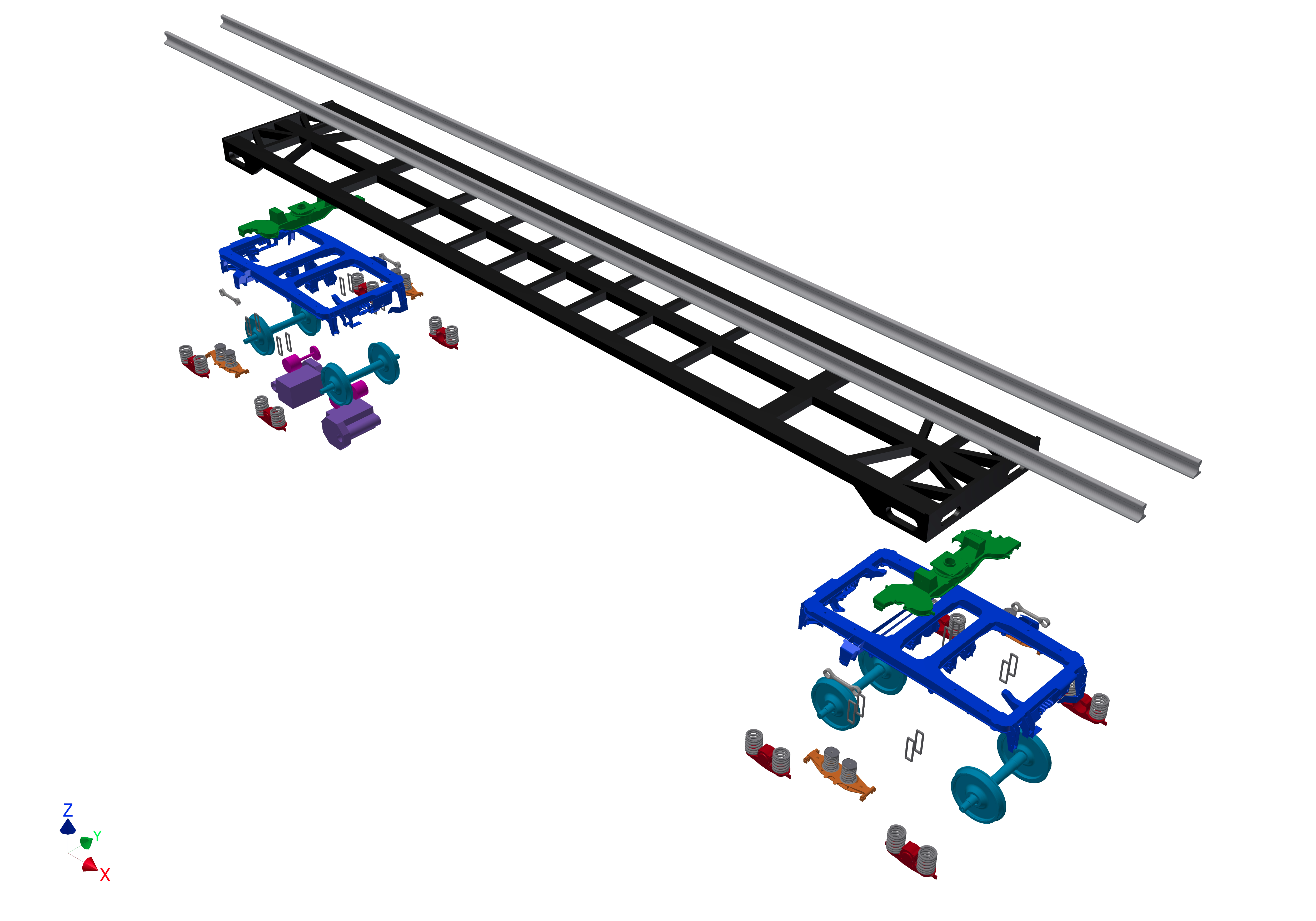}
\caption{Expanded View}
\label{fig:expanded_bodies}
\end{figure}

The \textit{FEVE 3000} \cite{feve3000} locomotive  multibody model developed in this work is depicted in Fig.~\ref{fig:locomotive} with an expanded view of the main parts shown in Fig.~\ref{fig:expanded_bodies}. 
%

The  $Vehicle\ Body$ (dark grey) is attached to the front and rear bogies thought two $Slider$ (green) bodies. The front bogie consists of a $Slider$ that rests on a couple of
$Suspender$ (orange) bodies hanging from the $Bogie\ Frame$ (dark blue), and two $Wheelset$ bodies (light blue) each of them with two $Axle\ Box$ (red) bodies. There is a couple of anti-yaw links (grey) between each $Suspender$ and the bogie frame. The rear bogie is identical to the front one but it includes two motors (one per each $Wheelset$). The  motor $Housing$ (mauve) rotates around the relative $Wheelset$ and is attached to the $Bogie\ Frame$ using a bushing. The motor includes a $Rotor$ (pink). The transmission of motion from the $Rotor$ to the $Wheelset$ is done by the way of a gear pair. 

%

The $Slider$ is connected by four identical spring-dampers to the Suspender part of the $Bogie\ Frame$. In a similar way, each $Axle\ Box$ is connected to the $Bogie\ Frame$ by two identical spring-dampers. The $Housing$ is also attached to the $Bogie\ Frame$ using a bushing. Compliance is considered in the gearing contacts.  Linear stiffness and damping is assumed for  spring-dampers, bushings and gear compliance.
%
%
%
Braking on the wheels and traction on the rotors is modeled considering externally applied torques.

The wheel-rail interaction model considers a fully three-dimensional rolling contact considering a single contact point per wheel. Normal contact is enforced through the use of constraints, while the tangential forces are determined based on the standard Kalker linear constitutive model.
Note that we consider generic wheel and rail profiles. The rails can present general irregularities along the track.

\subsection{Parametrization}
\subsubsection*{Multibody}
The $Vehicle\ Body$ is positioned relative to the track using absolute coordinates  (3 translations followed by 3 Euler  rotations).
Each $Slider$ is attached to the $Vehicle$ $Body$ by a revolute joint. A rotation relative to the $Vehicle\ Body$, in the vertical direction, is used to position the $Slider$.

To simplify the modeling, the effect of the anti-yaw bar is accounted for by removing the relative yaw motion between the $Bogie\ Frame$ and the $Slider$. With the same purpose,  the  $Suspender$ is considered fixed to the $Bogie\ Frame$. We use a vertical translation followed by two successive horizontal rotations (roll and tilt) to  position the $Bogie\ Frame$ relative to the $Slider$. Each $Axle\ Box$ is positioned fixed to the ``non-spinning wheelset'' frame (NSWHS), a frame that follows the relative $Wheelset$ but that does not spin with it. Each $Wheelset$ is positioned relative to the $Bogie\ Frame$ using a vertical translation and two horizontal rotations (roll and spin). Other relative degrees of freedom between these bodies are removed by the particular configuration of the spring-dampers. 
A rotation around the $Wheelset$ axis, relative to the NSWHS frame, is used to place each motor $Housing$. A rotation in the same direction, also relative to the NSWHS frame, is introduced to give the angular position of the $Rotor$ of each motor.

A total number of $60$ generalized coordinates, $\vq$, is used in this parametrization.

\subsubsection*{Contact}

The rails and wheel surfaces are described using cubic splines \cite{pombo2003general}, 

\begin{equation}\begin{aligned}
f^*(u^*) = (((a^* (u^*-u^*_{bp}) + b^* ) (u^*-u^*_{bp}) + c^*)  (u^*-u^*_{bp}) +d^*),
\label{eq:spline}
\end{aligned} \end{equation}
defined based on a set of control points that approximate their geometry. 


\begin{figure}
\centering
 \includegraphics[width=\textwidth]{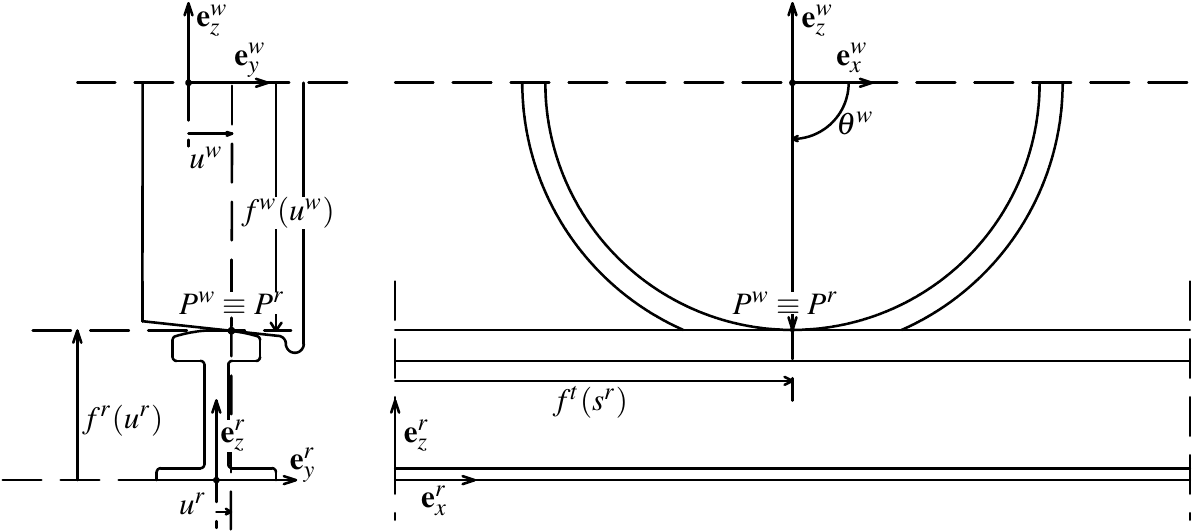}
\caption{Surfaces and railway parametrization}
\label{fig:surfaces}
\end{figure}

Figure \ref{fig:surfaces}  schematically shows the parametrization for a wheel-rail pair. The wheel and rail profiles are given respectively by $f^w(u^w)$ and $f^r(u^r)$, while the shape of the center line of the base of the rail along the track is given by $f_x^t(s^r)$, $f_y^t(s^r)$ and $f_z^t(s^r)$. Wheel surfaces are supposed to have cylindrical symmetry. Not represented in the figure is the spline used to represent the camber of the rail $f_\theta^t(s^r)$. 

For the contact point at each wheel-rail pair the parameters $\theta^w, u^w$ and $s^r, u^r$  play the role of ``generalized coordinates''  used to position the contact points, $P^w$ and $P^r$, respectively onto the wheel and rail surfaces. When the contact is materialized, the point $P^w$ 
and the point $P^r$ -defined as an arbitrary points in the surfaces of the wheel and rail respectively- are coincident.


The position of $P^w$ is given relative to a reference point in the wheel axis, $O^w$, as
\begin{equation}\begin{aligned}
\mathbf{r}_{O^w}^{P^w}(u^w,\theta^w) =  f^w(u^w) \cos(\theta^w) \mathbf{e}^w_x + u^w \mathbf{e}^w_y - f^w(u^w)\sin(\theta^w) \mathbf{e}^w_z.
\label{eq:OwPw}
\end{aligned} \end{equation}
For numerical reasons, the position of this point is given relative to a NSWHS base, $\mathbf{e}^w_x,\mathbf{e}^w_y ,\mathbf{e}^w_z$.
Analogously, the position of $P^r$ is given relative to a reference point in ground or track reference, $O^r$
\begin{equation}\begin{aligned}
\mathbf{r}_{O^r}^{P^r}(u^r,s^r) =  f_x^t(s^r)\mathbf{e}^r_x + f_y^t(s^r) \mathbf{e}^r_y + f_z^t(s^r) \mathbf{e}^r_z
+ u^r \hat{\mathbf{c}} + f^r(u^r) \hat{\mathbf{n}}
\end{aligned} \end{equation}
The base $\mathbf{e}^r_x,\mathbf{e}^r_y ,\mathbf{e}^r_z$  is fixed at the ground. Defining $\hat{\mathbf{t}}$ as a unit vector tangent to the center line of the rail base, $\hat{\mathbf{c}}$ is defined as a unit vector perpendicular to $\hat{\mathbf{t}}$ with an angle $f_\theta^t(s^r)$ with the ground measured in the positive direction of $\hat{\mathbf{t}}$. $\hat{\mathbf{n}}$ completes the base so that it is dexterous. 

 

The whole set of $8\times 4=32$ generalized coordinates required to position the $i=1,...,8$ contact points used in the analyzed example is referred as $\mathbf{s}=[...,\mathbf{s}_i\T,...]\T$. Where $\mathbf{s}_i=[\theta^w_i, u^w_i,s^r_i, u^r_i]\T$ is set of coordinates required to position points $P^w_i$ and $P^r_i$ at the $i$-th contact point. 

From the symbolic modeling point of view, functions $f^*(u^*)$ are the are modeled as if a single $3^{rd}$ order polynomial completely represents the whole profiles. At ``show-time'', the coefficients $a^*$, $b^*$, $c^*$, $d^*$ and break-points $u^*_{bp}$ are updated depending on the position of the contact point.

As commented before, for the purposes of this paper we only consider a single point of contact. Note, that
 flange  contact rarely occurs when the train runs along straight tracks or huge radii curved tracks unless train velocity is close to its critical speed \cite{andersson1999rail}. Nevertheless the parametrization proposed here is compatible
 some multiple-point-of-contact approaches \cite{andersson1999rail}.

\subsection{Constraint equations}\label{subsec:ConstrEqus}

 The only constrains that are present in the analyzed problem are those related to the contact points between wheel an rail.
 At a given contact, it should be enforced that the points $P^w$ and $P^r$ are coincident and that the surfaces at these points are tangent.
 
 Defining the tangent and normal vectors to the wheel at point $P^w$ as $\mathbf{t}_x^w$, $\mathbf{t}_y^w$, $\mathbf{n}^w$, and
 the tangent and normal vectors to the rail at point $P^r$ as $\mathbf{t}_x^r$, $\mathbf{t}_y^r$, $\mathbf{n}^r$. These conditions can be written
  \cite{Shabana2005} as: 
 \begin{equation}
\vphin (\vq,\vs)= \mathbf{n}^r \cdot \mathbf{r}^{P^w}_{P^r} = \vzero
\end{equation}
and
\begin{equation}
\vphid (\vq,\vs)= 
\left[
 \begin{array}{c}
  \mathbf{t}_x^r\cdot\mathbf{r}^{P^w}_{P^r}\\
  \mathbf{t}_y^r\cdot\mathbf{r}^{P^w}_{P^r}\\
  \mathbf{t}_x^w\cdot\mathbf{n}^r\\
  \mathbf{t}_y^w\cdot\mathbf{n}^r
 \end{array}
 \right]
 =
\vzero,
\end{equation}
 where $\vphin$ is the so called normal constraint, and $\vphid$ are the so called tangent constraints.
For each contact point, the tangent and normal vectors can be defined as:


\begin{eqnarray}
\mathbf{t}^r_x&=\dfrac{{\partial\mathbf{r}_{O^r}^{P^r}}/{\partial s^r}}{\left|{\partial\mathbf{r}_{O^r}^{P^r}}/{\partial s^r}\right|} 
, ~ ~
\mathbf{t}^r_y=\dfrac{{\partial\mathbf{r}_{O^r}^{P^r}}/{\partial u^r}}{\left|{\partial\mathbf{r}_{O^r}^{P^r}}/{\partial u^r}\right|} 
\text{, and   }  ~ ~ 
\mathbf{n}^r= \mathbf{t}^r_x \times \mathbf{t}^r_y\\
\mathbf{t}^w_x&=\dfrac{{\partial\mathbf{r}_{O^w}^{P^wr}}/{\partial \theta^w}}{\left|{\partial\mathbf{r}_{O^w}^{P^w}}/{\partial \theta^w}\right|} 
, ~ ~
\mathbf{t}^w_y=\dfrac{{\partial\mathbf{r}_{O^w}^{P^w}}/{\partial u^w}}{\left|{\partial\mathbf{r}_{O^w}^{P^w}}/{\partial u^w}\right|} 
\text{, and   }  ~ ~ 
\mathbf{n}^w= \mathbf{t}^w_x \times \mathbf{t}^w_y.
\end{eqnarray}
These vectors and constraint equations can easily be defined  using the symbolic procedures previously discussed. It will be seen that
this symbolic implementation will be very efficient as well.

Now, we use the subindex $i=1,...,8$ to refer to the constraint equations relative to each of the $8$ contact points. The set of all the normal constraints is referred as $\vphin (\vq,\vs)=[...,\vphin_i (\vq,\vs),...]\T$. Analogously,
the set of all the tangent constraints is referred as $\vphid (\vq,\vs)=[...,\vphid_i (\vq,\vs),...]\T$.


%
%
%
%
%

\subsection{Dynamic equations}
As commented previously the dynamic equations are obtained based on the direct application of the principle of virtual power. Using the vector $[\vq,\vs]$ as the set of generalized coordinates, the  mass matrix $\mathbf{M}$ is obtained by differentiation of the equations motion with respect to the the generalized accelerations and the generalized force $\boldsymbol{\delta}$ vector is obtained by substitution of the generalized accelerations by zero in the equations of motion. These equations should be complemented by the second derivative of the constraint equations to have a determined set equations
\begin{eqnarray}
\ddot{\vphin}(\vq,\vs)  &= \vzero  \label{eq:phin}\\
\ddot{\vphid}(\vq,\vs)  &= \vzero  \label{eq:phit},
\end{eqnarray}

 This set of equations shows the following structure:
\begin{align}
\begin{bmatrix} 
\mM_{\vq\vq}(\vq,\vdq)  & \vzero & \mdPhindqT(\vq,\vs) & \mdPhiddqT(\vq,\vs) \\
\vzero &\vzero & \vzero & \mdPhiddsT(\vq,\vs) \\
\mdPhindq(\vq,\vs)  & \vzero & \vzero &\vzero \\
\mdPhiddq(\vq,\vs)  & \mdPhidds(\vq,\vs) & \vzero &\vzero
\end{bmatrix}
\begin{bmatrix} \vddq \\ \vdds \\ \vlambda^n  \\ \vlambda^d \end{bmatrix} 
=
\begin{bmatrix} \vdelta_{\vq}(\vq,\vdq) \\ \vzero \\ \vgamman(\vq,\vs,\vdq,\vds) \\ \vgamma^d(\vq,\vs,\vdq,\vds) \end{bmatrix},
\end{align}
where $\mM_{\vq\vq}$ and $\vdelta_{\vq}$ are the blocks of the mass matrix  $\mathbf{M}$ and vector $\vdelta$ related to the set of coordinates $\vq$. 
In reference \cite{Shabana2005}, the authors refer this formulations as the Augmented Contact Constraint Formulation (ACCF).

\begin{figure}[h]
\centering
 \includegraphics[height=80mm]{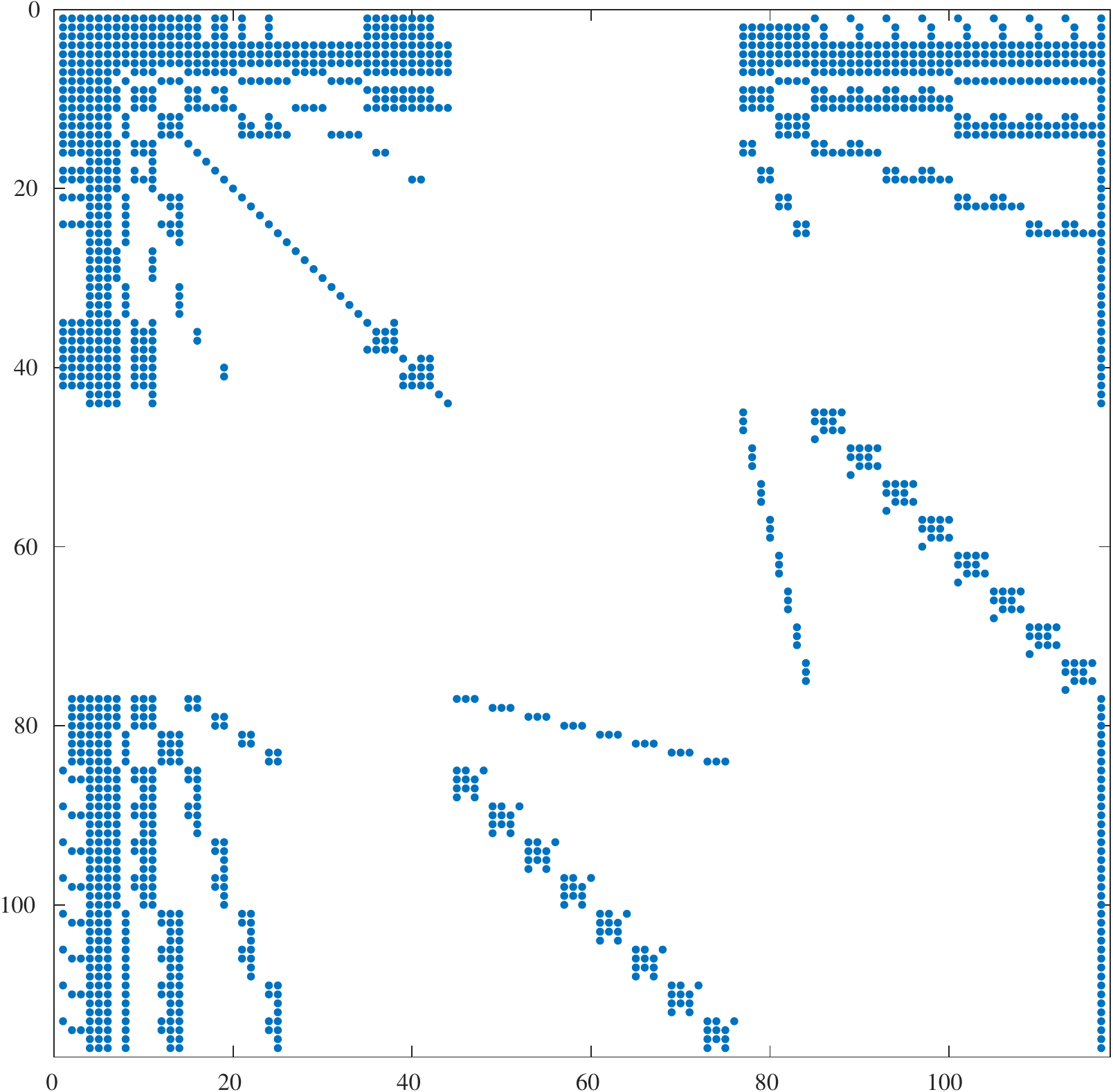}
\caption{Dynamic model structure $\left[[ \mM,\mdPhidqT ;\mdPhidq,\vzero],[\vdelta;\vgamma ]\right]$}
\label{fig1}
\end{figure}

\begin{figure}[h]
\centering
 \includegraphics[height=32mm]{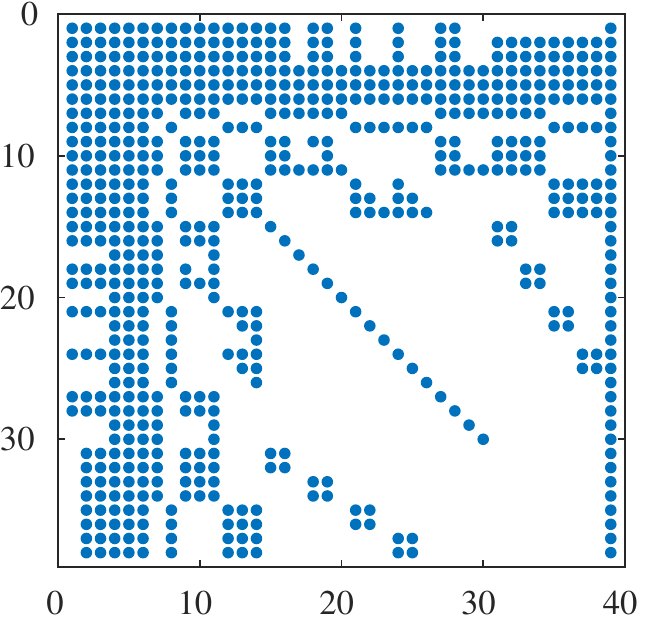}
\caption{Dynamic model structure $\left[[ \mM_{\vq\vq},\mdPhindqT ;\mdPhindq,\vzero],[\vdelta_{\vq};\vgamman]\right]$}
\label{fig2}
\end{figure}

The particular structure of the dynamic equations for the problem analyzed can appreciated on Fig.~\ref{fig1}. There, the nonzero entries for matrix $\left[[ \mM,\mdPhidqT ;\mdPhidq,\vzero],[\vdelta;\vgamma ]\right]$ are shown as dots. It should be noted that, $\mdPhinds$ is zero numerically, even if symbolically there can be seen few nonzero expressions.
As accelerations $\vdds$ are not needed, the previous system of equations can be reduced to
\begin{equation}
\begin{bmatrix} 
\mM_{\vq\vq}(\vq,\vdq)   & \mdPhindqT(\vq,\vs)  \\
\mdPhindq(\vq,\vs)  & \vzero
\end{bmatrix}
\begin{bmatrix} \vddq \\ \vlambda^n \end{bmatrix} \label{eq:MphiTPhiZero4}
=
\begin{bmatrix} \vdelta_{\vq}(\vq,\vdq) \\ \vgamman(\vq,\vs,\vdq,\vds) \end{bmatrix}, 
\end{equation}
The reduced structure of the dynamic equations can appreciated on Fig.~\ref{fig2}. Obviously, standard linear solution procedures are going to perform much more efficiently in this case.
There, the nonzero entries for matrix $\left[[ \mM_{\vq\vq},\mdPhindqT ;\mdPhindq,\vzero],[\vdelta_{\vq};\vgamman]\right]$ are shown as dots.
In reference \cite{Shabana2005}, the authors refer this formulation as the Embedded Contact Constraint Formulation (ECCF) \cite{Shabana2005}.

Constraint stabilization is performed based on projection on the coordinate and velocity manifolds. That requires the solution of the following equations
\begin{eqnarray}
\vphin(\vq,\vs)  &= \vzero  \label{eq:phin}\\
\vphid(\vq,\vs)  &= \vzero  \label{eq:phit},
\end{eqnarray}
at the coordinate level, and of
\begin{eqnarray}
\mdPhindq(\vq,\vs) \vdq &= \vbeta^n(\vq,\vs) = \mathbf{0} \label{eq:dqproj} \\ 
\mdPhiddq(\vq,\vs) \vdq + \mdPhidds(\vq,\vs) \vds &= \vbeta^d(\vq,\vs) = \mathbf{0} \label{eq:dsproj}
\end{eqnarray}
at the velocity level. Note that there are no rheonomous equations in the problem analyzed, an therefore $\vbeta^n(\vq,\vs) $ and $\vbeta^d(\vq,\vs)$ are $\mathbf{0}$. For example, we frequently use coordinate partitioning \cite{Wehage1982,Haug1989} into dependent and independent. Dependent coordinates and velocities are obtained in terms of the independent ones.

It should be remarked that in the ECCF context, $\vs$ and $\vds$ are better not considered standard generalized coordinates, but a set of auxiliary variables that must be known in order to solve equation system  (\ref{eq:MphiTPhiZero4}). After the integration, Eq.~(\ref{eq:phit}) and Eq.~(\ref{eq:dsproj}) can be used to
obtain the auxiliar variables $\vs$ and $\vds$ in terms of $\vq$ and $\vdq$. In this context the position and velocity projection would be performed after this step, based
on Eq.~(\ref{eq:phin}) and Eq.~(\ref{eq:dqproj}), respectively. These inertia-less coordinates find different names in the literature, such as ``surface parameters'' \cite{shabana2004development,Shabana2005,pombo2007new,auciello2009dynamic,arnold2004simulation}, ``non-generalized coordinates'' \cite{shabana2001augmented} or  ``auxiliary variables'' \cite{fisette2000multibody}, but their main feature is that they do not participate in the system dynamics.

\subsection{Contact model.}

In the context of railway dynamic simulation is very important to correctly determine the values of the creep forces between the wheel and the rail. To that end, in this work the well known linear contact theory of Kalker  \cite{kalker1967rolling} is used. This theory requires the determination of several data: location of the contact point, the creepages, normal loads at this contact point, wheel and rail surface curvatures at the contact point, tangent and normal vectors at the contact patch. This computations are
big and must be done for every contact. To speed up computations, several authors \cite{bozzone2011lookup,sugiyama2011wheel,escalona2015modeling} propose the use of pre-calculated look-up tables to determine the required data. This procedure is tedious and usually requires to introduce some modeling simplifications. We propose to compute these quantities without simplifications, on line, based on functions exported using the proposed symbolic methods. This is a more simple and general procedure to apply. The results will confirm that this is very fast procedure.

\subsubsection*{Contact patch geometry determination.}

Based on classical Hertzian contact theory, the contact patch is a flat ellipse \cite{Iwnicki2006}. The semi-axes of this ellipse in the longitudinal and transversal directions, $a$ and $b$ respectively, are determined as follows:
\begin{equation}
 a=\left(\frac{3}{2}  \frac{1-\nu^2}{E}\frac{1}{A+B} N \right)^{\frac{1}{3}}m(\theta)~~~\text{and}~~~
 b=\left(\frac{3}{2}  \frac{1-\nu^2}{E}\frac{1}{A+B} N \right)^{\frac{1}{3}}n(\theta).
\end{equation}
In these expressions, $N$ is the normal contact force acting on the wheel, $E$ is the Young’s modulus, $\nu$ the Poisson’s ratio. $m(\theta)$ and $n(\theta)$ are adimensional functions proposed by Hertz. We use on-line interpolation in table $4.1$ in Ref.~\cite{Iwnicki2006} to evaluate these functions.  $\theta=\cos^{-1}{\left(\frac{\left|A-B\right|}{A+B}\right)}$, where  $A$ and $B$ are determined as
\begin{equation}
A=\frac{1}{2}\left(\frac{1}{R^{w}_{x}}+\frac{1}{R^{r}_{x}}\right)~~~\text{and}~~~B=\frac{1}{2}\left(\frac{1}{R^{w}_{y}}+\frac{1}{R^{r}_{y}}\right).
\end{equation}
$R^{w}_{x}$, $R^{w}_{y}$, $R^{r}_{x}$, $R^{r}_{y}$ are the curvature radii of wheel and rail surfaces at the contact point. For the case studied these are computed as:
\begin{equation}
\begin{array}{cc}
R^{w}_{x}= \dfrac{\sqrt{(\mathbf{r}_{O^w}^{P^w}\mathbf{e}_{x}^{w})^2+(\mathbf{r}_{O^w}^{P^w}\mathbf{e}_{y}^{w})^2}}{\sqrt{1-\left(\mathbf{n}^r \mathbf{e}^w_y\right)^2}}~~~~~\footnotemark
& 
~~~~~R^{w}_{y}=\left|\dfrac{(1+\frac{\partial f^w}{\partial{u_w}})^{3/2}}{\frac{\partial^2 f^w}{{\partial u_w}^2}}\right|
\\
R^{r}_{y}=\left|\dfrac{(1+\frac{\partial f^r}{\partial{u_r}})^{3/2}}{\frac{\partial^2 f^r}{{\partial u_r}^2}}\right| 
&
R^{r}_{x}=\infty
\end{array}
\end{equation}
\footnotetext{Distance from the rotation axis to the contact point along the normal at contact point.}

These curvature radii are obtained and exported based on the symbolic methods presented preciously in this paper. The normal force $N$ is the Lagrange multiplier associated with the normal constraint of the relative contact point. It is obtained directly from the solution of the dynamic system of
equations. To keep the dynamic problem linear, avoiding a nonlinear iteration, the normal force used is the one obtained in the previous integration step.

\subsubsection*{Creep forces and moments}
From the modeling perspective creep forces and moments are considered external actions, so their effect is contained in vector $\boldsymbol{\delta}_{\vq}$, along with all the contributions due to  other inertial constitutive and external forces. 
For each wheel the these forces and moments are symbolically expressed as
\begin{equation}
 \mathbf{f}=f_x~\mathbf{t}^r_x+f_y~\mathbf{t}^r_y ~~~~\text{and}~~~~ \mathbf{m}=m_z~\mathbf{n}^r,
\end{equation}
where $f_x$, $f_y$ and $m_z$ are symbols. These represents the tangent contact force and spin contact moment acting on the wheel at the contact point $P^w$. They are defined at the contact base  $\mathbf{t}^r_x$,$\mathbf{t}^r_y$, $\mathbf{n}^r$.
The components of these vectors are numerically computed as 
\begin{equation}
\left[
 \begin{array}{c}
  f_x\\
  f_y\\
  m_z
 \end{array}
\right]=-G
\left[
 \begin{array}{ccc}
  a b ~c_{11}&0&0\\
  0& a b ~c_{22}& \sqrt{ab}~c_{23}\\
  0&-\sqrt{ab}~c_{23}& (ab)^2~c_{33}
 \end{array}
 \right]
\left[
 \begin{array}{c}
  \xi_x\\
  \xi_y\\
  \varphi_z
 \end{array}
 \right]
\label{eq:kalker1}
\end{equation}
\cite{Iwnicki2006}. Parameter $G$ is the material shear modulus and $c_{ij}(a/b,\nu)$ are the coefficients determined by Kalker, tabulated in Ref.~\cite{kalker1968tangential}. We use on-line interpolation in these tables. The creepages $\xi_x$, $\xi_y$ and  $\varphi$ are defined as:
%
%
\begin{eqnarray}
\xi_x = \frac{\mathbf{v}_{Gr.}^{P^w}}{\frac{1}{2}(\left|\mathbf{v}_{Gr.}^{O^w}\right|+\left|\boldsymbol{\omega}_{Gr.}^{w} \wedge \mathbf{r}_{O^w}^{P^w}\right|)}~\mathbf{t}^r_x \\
\xi_y = \frac{\mathbf{v}_{Gr.}^{P^w}}{\frac{1}{2}(\left|\mathbf{v}_{Gr.}^{O^w}\right|+\left|\boldsymbol{\omega}_{Gr.}^{w} \wedge \mathbf{r}_{O^w}^{P^w}\right|)}~\mathbf{t}^r_y  \\
\varphi_z = \frac{\boldsymbol{\omega}_{Gr.}^{w}}{\frac{1}{2}(\left|\mathbf{v}_{Gr.}^{O^w}\right|+\left|\boldsymbol{\omega}_{Gr.}^{w} \wedge \mathbf{r}_{O^w}^{P^w}\right|)}~\mathbf{n}^r
\end{eqnarray}
where $\mathbf{v}_{Gr.}^{P^w}$ in the velocity with respect to the ground ($Gr.$) of the $P^w$ contact point when moves ``attached'' to the wheel and $\boldsymbol{\omega}_{Gr.}^{w}$ is the wheel-set angular velocity with respect to the ground. $\mathbf{v}_{Gr.}^{O^w}$ in the velocity with respect to the ground of the center of the wheel-set and $\mathbf{r}_{O^w}^{P^w}$ is the position vector form $O^w$ to $P^w$ as shown in Eq.~(\ref{eq:OwPw}). The creepages are determined numerically based on exported functions for the  numerators and denominators of these expressions. In this way division by zero can  be dealt within the numerical solver.

All the symbolic functions required for the implementation of the Kalker model for all the different contact points, referred previously, are computed in a single function call. In this way recycling of atoms is maximized.

\section{Numerical integration}
\label{sec3}
For our case study the Linear Kalker model used to model the contact forces has been a mayor source of problems.
To some extent, these forces can be considered viscous friction forces with a
huge value for the equivalent viscous constant, making the system dynamics stiff. In this context, the use of an explicit integration scheme
is going to require a very small time step, spoiling real-time performance.

In order to solve this problem we have devised an Implicit-Explicit (IMEX) \cite{ascher1995implicit}  integration schema, and adjusted it to overcome the problem associated with the contact forces  without penalizing the computational cost.
The use of these schemes is not new. It has appeared in the bibliography under other names: semi-implicit \cite{arnold2007linearly}, additive or combined methods \cite{fulton2004semi}, etc. These methods use different types of discretization for the different terms in the dynamic equations. Those terms that are not related to the stiff behavior of the equations are discretized  using a low-cost explicit scheme while, the stiff terms are discretized using an implicit scheme.

As commented before creep forces and moments are introduced in the model as external actions, and their contribution is embedded  in vector $\vdelta_{\vq}$. Lets make this contribution explicit, by splitting $\vdelta$  into two  Kalker ($K$) and non-Kalker ($NK$) contributions.
In this way, first equation in Eq.~(\ref{eq:MphiTPhiZero4}) can be rewritten as:

\begin{equation} 
\begin{bmatrix} 
\mM_{\vq\vq}   & \mdPhindqT  \\
\mdPhindq  & \vzero
\end{bmatrix}
\begin{bmatrix} \vddq \\ \vlambda^n \end{bmatrix} 
=
\begin{bmatrix} \vdelta^{K}_{\vq} + \vdelta^{NK}_{\vq} \\ \vgamman \end{bmatrix} 
\end{equation}

This new set of equations can be integrated using an IMEX method. The terms related to creep forces will be integrated using an Implicit scheme and the rest using an Explicit scheme.

Eq.~(\ref{eq:kalker1}) can be expressed as a typical viscous contribution
\begin{equation}
\mathbf{f}^K= -\frac{1}{V}
\mathbf{C}^K
\boldsymbol{\nu}
\label{eq:kalker_viscous}
\end{equation}
where $\mathbf{f}^K=[f_x,f_y, m_z]\T1$, $V=\frac{1}{2}(\left|\mathbf{v}_{Gr.}^{O^w}\right|+\left|\boldsymbol{\omega}_{Gr.}^{w} \wedge \mathbf{r}_{O^w}^{P^w}\right|)$, $\boldsymbol{\nu} = [\mathbf{v}_{Gr.}^{P^w} \mathbf{t}^r_x ~~ \mathbf{v}_{Gr.}^{P^w}\mathbf{t}^r_y ~~ \boldsymbol{\omega}_{Gr.}^{w}\mathbf{n}^r]\T$ and 

\begin{equation}
 \mathbf{C}^K=G
\left[
 \begin{array}{ccc}
  a b ~c_{11}&0&0\\
  0& a b ~c_{22}& \sqrt{ab}~c_{23}\\
  0&-\sqrt{ab}~c_{23}& (ab)^2~c_{33}
 \end{array}
 \right]
\end{equation}
Adding a subindex $i$ to refer to a particular contact point, the contribution $\vdelta^{K}$ can be obtained as:
\begin{align} 
\vdelta^{K}_{\vq} = \sum_{i=1}^8 \frac{\partial \vdelta^{K}_{\vq}}{\partial \mathbf{f_i}^{K}} \mathbf{f_i}^{K} =
- \sum_{i=1}^8 \frac{\partial \vdelta^{K}_{\vq}}{\partial \mathbf{f_i}^{K}} \frac{1}{V_i} \mathbf{C}^K_i
\frac{\partial \boldsymbol{\nu}_i}{\partial \vdq} \vdq = - \mathbf{C}^{K}_{\vq\vq} \vdq 
\end{align}
In order to determine matrix $ \mathbf{C}^{K}_{\vq\vq}$, we symbolically export matrices
\begin{equation}
                   \frac{\partial \vdelta^{K}_{\vq}}{\partial \mathbf{f_i}^{K}} \mathbf{C}_i^{K}\frac{\partial \boldsymbol{\nu}_i}{\partial \vdq}  ~,~~~i=1,\ldots,8,                                                                      
\end{equation}
and numerically assemble matrix as
\begin{equation}
 \mathbf{C}^{K}_{\vq\vq}= \sum_{i=1}^8 \frac{1}{V_i} ~~\frac{\partial \vdelta^{K}_{\vq}}{\partial \mathbf{f_i}^{K}}  \mathbf{C}^K_i
\frac{\partial \boldsymbol{\nu}_i}{\partial \vdq},
\end{equation}
where $V_i$ and $\mathbf{C}^K_i$ are determined using the same procedures described in the previous section.

The contribution  $\vdelta^{NK}$, can be obtained symbolically substituting by zero en $\vdelta_{\vq}$ the symbols associated to the external forces
$f_x,f_y, m_z$ for every contact point.

Now write the dynamic equation set can be expressed as follows:

\begin{align} 
\begin{bmatrix} 
\mM_{\vq\vq}   & \mdPhindqT  \\
\mdPhindq  & \vzero
\end{bmatrix}
\begin{bmatrix} \vddq \\ \vlambda^n \end{bmatrix} 
&=
\begin{bmatrix}  \vdelta^{NK}_{\vq}-\mathbf{C}^{K}_{\vq\vq}\vdq \\ \vgamman \end{bmatrix}
\label{eq:MphiTPhiZero5}
\end{align}
The IMEX integration procedure proposed follows directly from this equation.

Must be observed that in Kalker's Linear Theory when saturation occurs this method is also valid, because in this case force can be also written as the product of a constant matrix and the creepages. The numerical solver must handle with which matrix use at each moment.

\subsubsection*{Discretization}

The contribution $\mathbf{C^{K}}_{\vq\vq}\vdq$ is discretized using an implicit Euler. To that end it is evaluated at the next time step $t+\Delta t$,
\begin{equation}
 \mathbf{C}^{K}_{\vq\vq}\vdq_{t+\Delta t}
\end{equation}
An explicit Euler scheme for the remaining terms requires acceleration to be discretized as
\begin{equation}
 \vddq_{t+\Delta t}=\frac{\vdq_{t+\Delta t}-\vdq_{t}}{\Delta t}
\end{equation}
and $\vdelta^{K}_{\vq}$ to be evaluated at $t$.

\begin{figure}[h]
\centering
 \includegraphics[height=32mm]{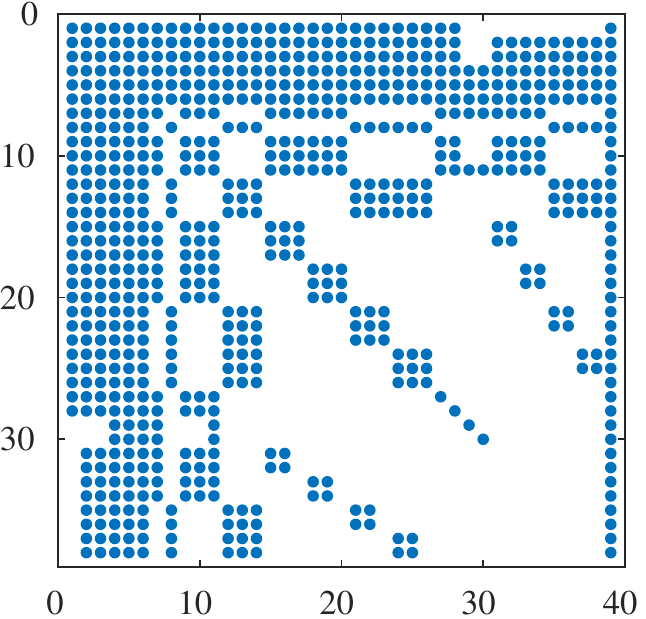}
\caption{Dynamic model structure $\left[[ \mM_{\vq\vq} + \mathbf{C}^{K}_{\vq\vq} \Delta t,\mdPhindqT \Delta t;\mdPhindq,\vzero],[ \vdelta^{NK}_{\vq} \Delta t + \mM_{\vq\vq} \vdq_{t};\vgamman \Delta t + \mdPhindq \vdq_{t}]\right]$}
\label{fig3}
\end{figure}

Substituting this into Eq.~(\ref{eq:MphiTPhiZero5}) the final discretization of the system
takes the form:
\begin{align} 
\begin{bmatrix} 
\mM_{\vq\vq} + \mathbf{C}^{K}_{\vq\vq} \Delta t &~ &\mdPhindqT  \Delta t\\
\mdPhindq  &~& \vzero
\end{bmatrix}
\begin{bmatrix} \vdq_{t+\Delta t} \\ \vlambda^n \end{bmatrix} 
&=
\begin{bmatrix}  \vdelta^{NK}_{\vq} \Delta t + \mM_{\vq\vq} \vdq_{t} \\ \vgamman \Delta t + \mdPhindq \vdq_{t} \end{bmatrix}
\label{eq:MphiTPhiZero6}
\end{align}
where all the functions are computed at time $t$.
Note that, to keep the equation solution linear, $\mathbf{C}^{K}$ is evaluated at $t$ instead of $t+\Delta t$.
The structure of this system of equations can be observed in Fig.~ \ref{fig3}. It is noticeable that the sparsity structure is very similar to the one seen in Fig.~\ref{fig2}.

This problem has the same mathematical structure that the standard full dynamic set, so it can be solved using the same procedures. We use coordinate partitioning \cite{Wehage1982,Haug1989}. This is a good performing strategy that is also used by other practitioners in the symbolic multibody field.
We use a $LU$ procedure with full pivoting on the non tangent constraint Jacobian in the generalized velocities $\vdq$, so we can choose the set of independent coordinates at each iteration step. This way, no conditions are enforced on the parameterization $\vq$ used.    
Thus using this IMEX  scheme comes for free, as the evaluation of the functions appearing in Eq.~ (\ref{eq:MphiTPhiZero6}) has the same complexity as the functions in Eqs.~ (\ref{eq:MphiTPhiZero5})  or (\ref{eq:MphiTPhiZero4}).

It should be noted that the matrix $\mM_{\vq\vq} + \mathbf{C}_{\vq\vq}^{K} \Delta t$ is not symmetric, so  $LU$ decomposition should be used in place of $LDL^{\T}$ incurring a small penalty in performance. 

The solution of this system will give the value of the generalized velocities at $t+\Delta t$, $\vdq_{t+\Delta t}$. To obtain the coordinates
at $\vq_{t+\Delta t}$ the following explicit mid-point rule is used:
\begin{equation}
 \vq_{t+\Delta t}=\vq_{t}+\frac{\vdq_{t+\Delta t}+\vdq_{t}}{2}{\Delta t}.
\end{equation}
Note that it is second order and comes at no cost.

Next coordinate projection is performed. First Eq.~(\ref{eq:phit}) is used to obtain the contact coordinates $\vs_{t+\Delta t}$ in terms of the $\vq_{t+\Delta t}$. As $\vq_{t+\Delta t}$ is accurate to second order, this procedure gives a  error of the same order. To this end the following iterative Newton-Raphson procedure is used:
\begin{equation}
%
   \mdPhidds(\vq,\vs) (\vs_{k+1}-\vs_k) =  -\vphid(\vq,\vs)  \label{eq:NRs}
\end{equation}
this usually involves a single iteration\footnote{This is related to the NSWHS frame used to define the contact point in the wheel.}. After the update of $\vs$, Eq.~(\ref{eq:phin}) is solved for $\vq$ using the same iterative procedure:
\begin{equation}
%
   \mdPhindq(\vq,\vs) (\vq_{k+1}-\vq_k) =  -\vphin(\vq,\vs)  \label{eq:NRq}
\end{equation}
This procedure usually converges in a single iteration. Note that $\vq_{t+\Delta t}$ is accurate to second order after the integration step. Note that the LU decomposition of the previous Jacobians, $\mdPhindq(\vq,\vs)$ and $\mdPhiddq(\vq,\vs)$, is known as they have computed at the previous velocity projection step (described latter). So the Jacobian and its decomposition is not updated in this step.

In the velocity projection step, first Eq.~(\ref{eq:dqproj}) is solved  for $\vdq$. To that end, the Jacobian $\mdPhindq(\vq,\vs)$ and its $LU$ decomposition are  updated. Then, Eq.~(\ref{eq:dsproj}) is solved for $\vds$. To that end, the Jacobians $\mdPhiddq(\vq,\vs)$  and  $\mdPhidds(\vq,\vs)$ are updated
and the $LU$ decomposition of $\mdPhidds(\vq,\vs)$ is computed.

\begin{figure}[h] 
\centering
\includegraphics[width=\textwidth]{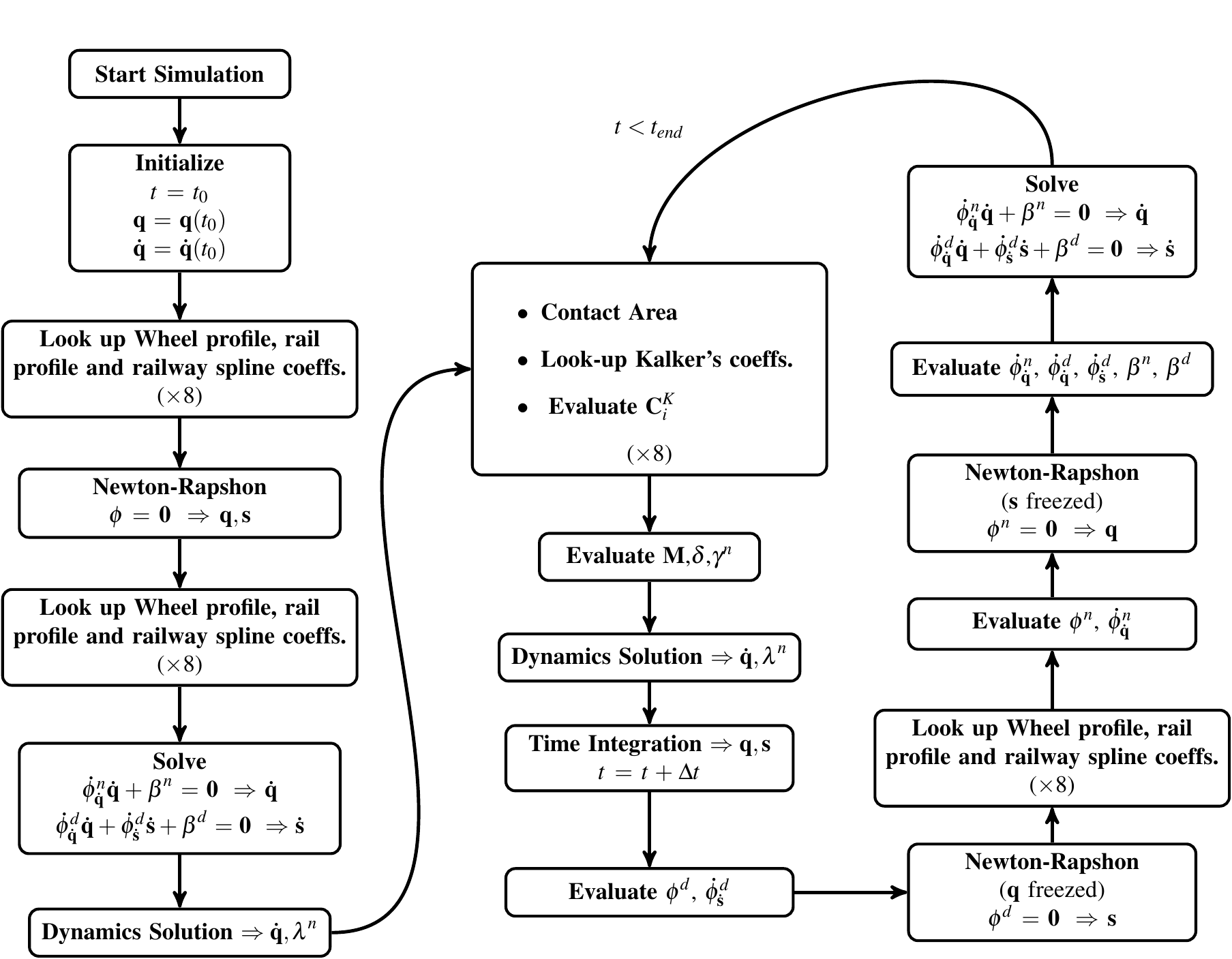}
\caption{Integration procedure}
\label{fig:integration}
\end{figure}

In Fig.~\ref{fig:integration}, a schematic representation of the integration procedure described here is presented.
To get a more clear picture, the steps related to the determination of the creep forces has been represented.

\section{Results}
\label{sec4}

\subsubsection*{Simulation description}

The track used in the simulation starts and ends with two straight and
parallel segments running in the $x$ direction and separated $50 ~\text{m}$.
Both stretches are joined by a symmetric and smooth double transition curve  $270 ~\text{m}$ long in direction $x$.
On top of the defined geometry, two harmonic vertical irregularities with an amplitude of $10~\text{mm}$ are added. These irregularities  are defined
using a sine wave that runs in direction of $x$ with a wave length of $10~\text{m}$. Right and left rail
irregularities present a phase difference of $\pi/2$. As commented earlier, third order splines are used to discretize the whole track, including the irregularity.

The simulation starts with an initial forward speed of $23.7~\text{m/s}$ with the $Vehicle$ $Body$ centered at $x=0 ~\text{m}$ and with a lateral misalignment of $5~\text{mm}$ with respect to to the track center.
Vehicle motors are actuated with a constant $200~\text{Nm}$ torque.

\subsubsection*{Computational results}

\begin{figure}[h]
  \centering
  \includegraphics[width=0.99\textwidth]{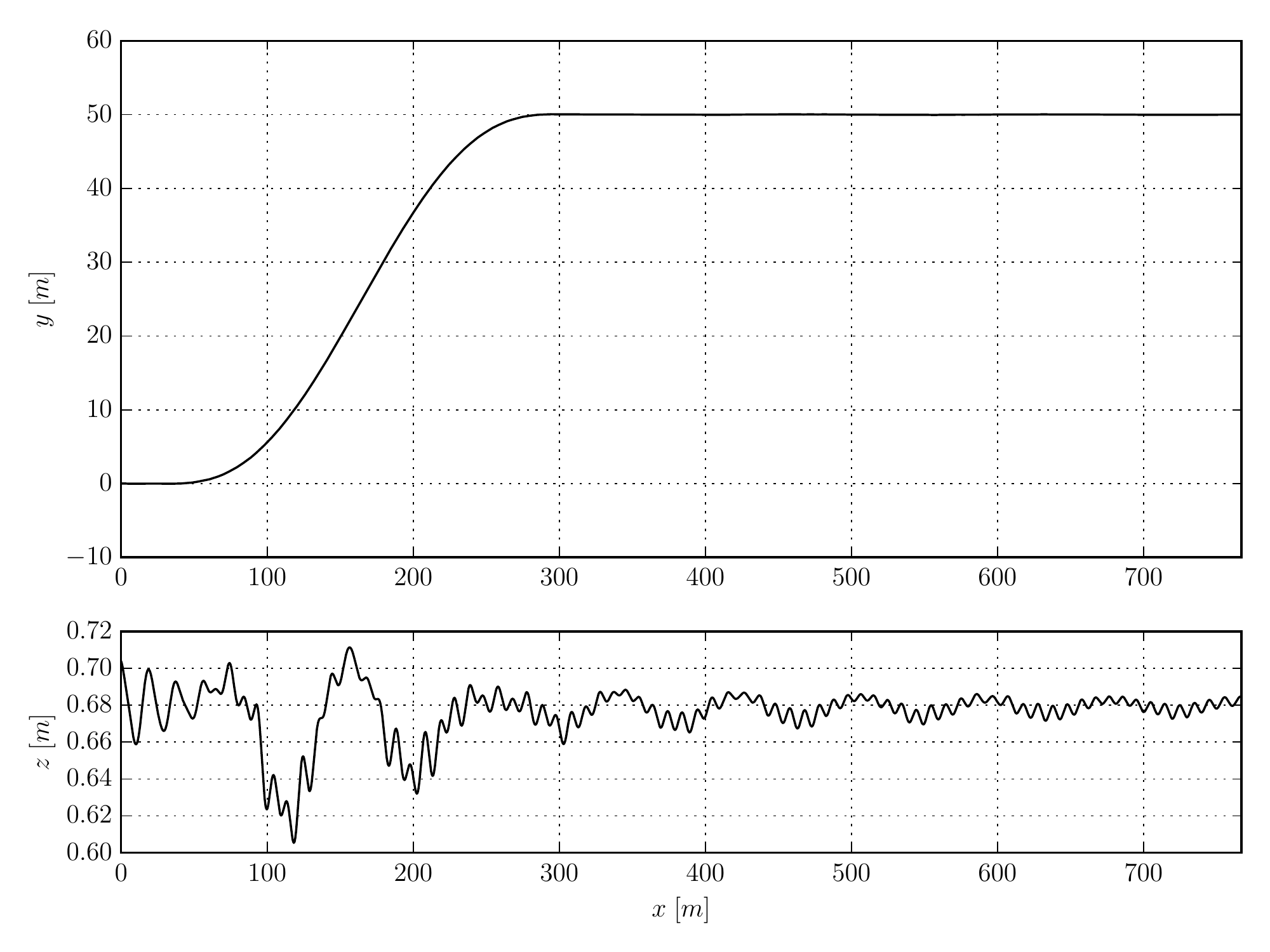}
  \caption{Trajectories followed by the main body}
  \label{fig:xzmainbody2}
\end{figure}

Fig.~\ref{fig:xzmainbody2} shows the trajectory followed by the $Vehicle\ Body$ center.
Note that the the given initial state is not in dynamic equilibrium and therefore, the oscillations at the beginning 
of the simulation are in part due to this. This is related to the sudden application of torque at the simulation start.
By the time that the vehicle center enters the track, the oscillations seen are no longer related to the initial condition.

\begin{figure}[h]
  \centering
  \includegraphics[width=0.99\textwidth]{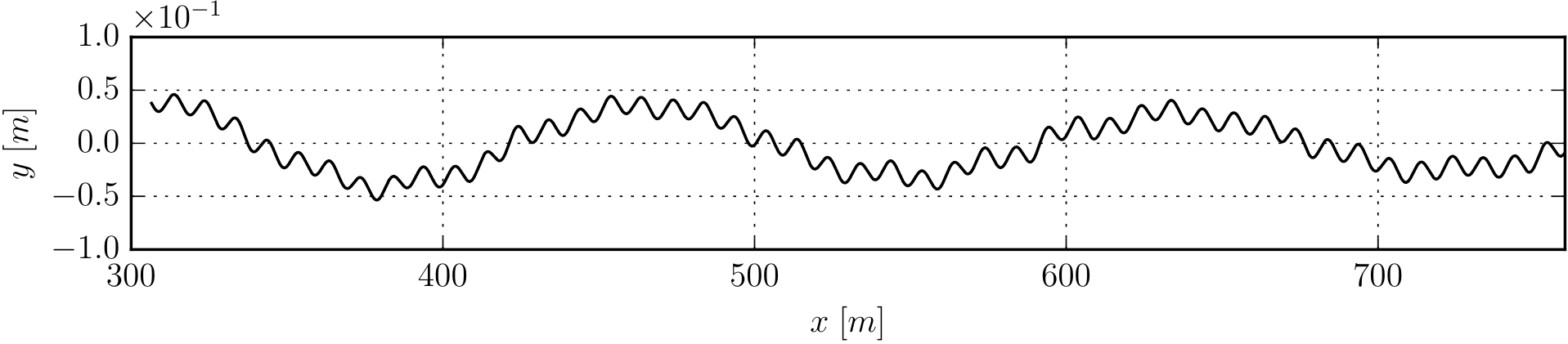}
  \caption{Hunting oscillation in detail.}
  \label{fig:hunting}  
\end{figure}

Fig.~\ref{fig:hunting} shows a zoom of the first graph in Fig.~\ref{fig:xzmainbody2}. This is done to make the oscillations
in that plane visible. The zone in which the vehicle  exits the second curve is shown. 
Two different oscillations can be seen. Two oscillations are clearly distinguishable: The hunting oscillation is the one with the largest wave length, while the shorter one is related
to the irregularities of the track. 

\begin{figure}[h]
  \centering
  \includegraphics[width=0.99\textwidth]{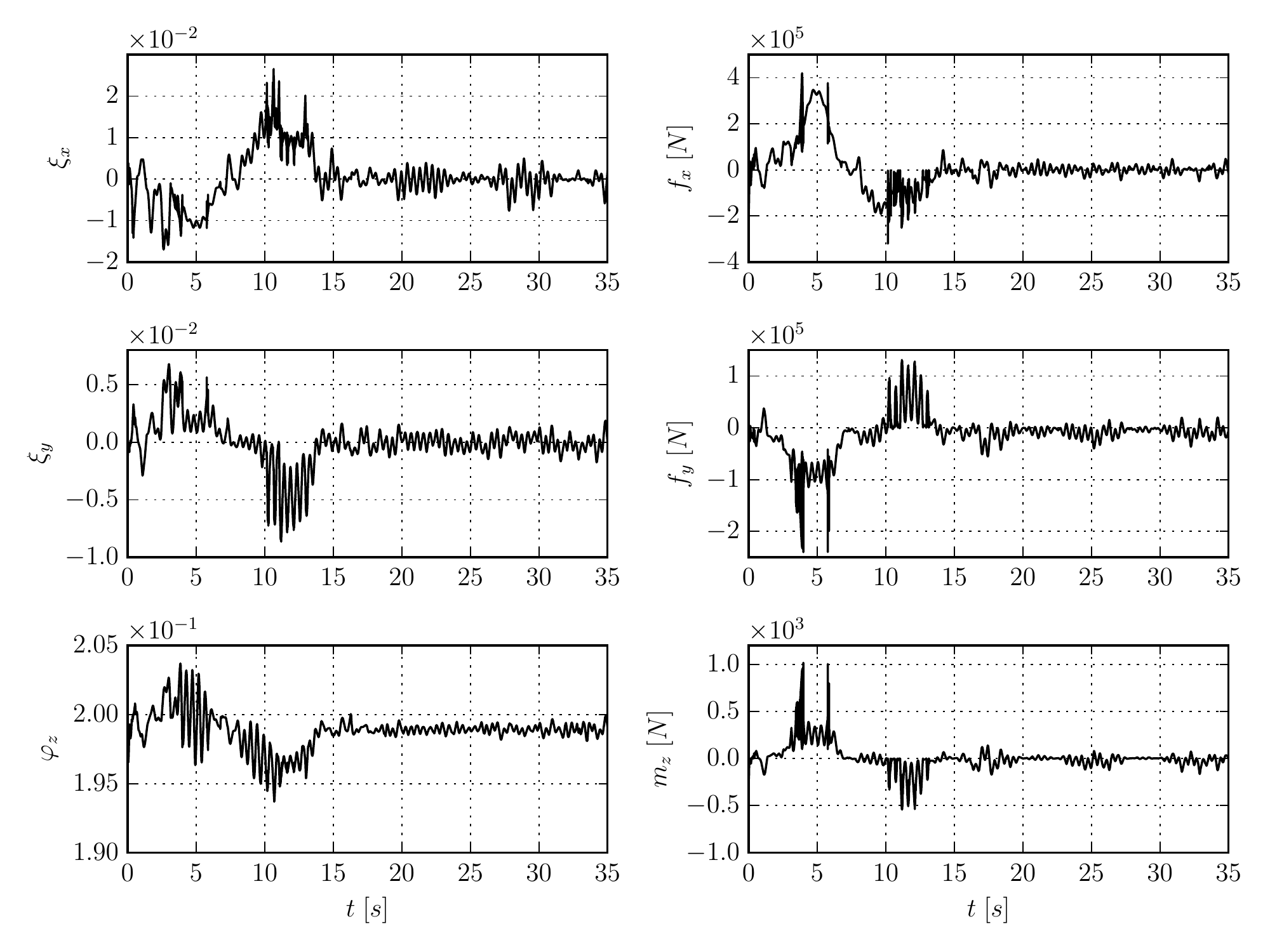}
  \caption{Creep velocities and  forces the Rear Bogie, Front-Right wheelset }
  \label{fig:kalkerforces}
\end{figure}

Creep velocities and creep forces and moments are presented in Fig.~\ref{fig:kalkerforces}. 
It can be observed that creep velocities are higher when the vehicle is at the middle of the curved tracks
($t\approx  5~s$ and $t\approx  10~s$). The same behavior is seen for the forces and moments. Small fluctuations
on the creepages and forces in the second straight track ($t> 15~s$) are due to the vertical irregularities.

Using the proposed IMEX integrator with a $\Delta t = 1~\text{ms}$ a stable integration is achieved by a generous margin.
In the same conditions, using an explicit Euler for velocities and the explicit mid-point rule for accelerations time steps smaller than
$10^{-4}~\text{ms}$, not compatible with a real-time performance, are required. Implicit trapezoidal rule has also been used with showing a stable
behavior around $\Delta t \approx 1~\text{ms}$ with not such a generous margin.

Using the IMEX  Euler method a fine grained analysis of the  computation times required by the different steps of the proposed algorithm is done.
The results are given in Table~ \ref{table:times}. A  seven years old \textit{Intel Core vPro i5 @ 3500}{MHz} has been used for the test.
From this data it can be seen that it takes $256 ~\mu \text{s}$ of CPU time to complete one integration step. That is, \textit{hard} real-time performance
is achieved by a wide margin using the proposed procedures. In comparison, using the trapezoidal rule \textit{soft} real-time performance can be achieved by a short margin.

\begin{table}[]
\centering
\caption{Results per time step ($1~\text{ms}$)}
\label{table:times}
\begin{tabular}{lrr}
\hline
Task                                                                                  & CPU Time $\mu \text{s}$  \\ 
\hline
Contact Area, Look up Kalker coeffs. and evaluate $\frac{\partial \vdelta^{K}_{\vq}}{\partial \mathbf{f_i}^{K}} \mathbf{C}_i^{K}\frac{\partial \boldsymbol{\nu}_i}{\partial \vdq}$ ($\times 8$)              & 30  \\
Look up Wheel profile, rail profile and  railway spline coeffs. ($\times 8$)          & 1  \\
Evaluate $\mM$,$\vdelta$ and $\vgamman$                                               & 55  \\
Dynamics Solution $\Rightarrow \vdq, \vlambda$                                        & 103  \\
Time Integration  $\Rightarrow \vq$                                                   & 1  \\
Evaluate   $\vphin$, $\vphid$, $\mdPhindq$, $\mdPhiddq$, $\mdPhidds$ and $\vbeta^n$   & 29  \\
Projection Solution $\Rightarrow \vq, \vdq, \vs, \vds$                                & 37  \\ 
\hline
Total Time                                                                            & 256  \\ 
\hline
\end{tabular}
\end{table}

\begin{figure}[h]
  \centering
  \includegraphics[width=0.99\textwidth]{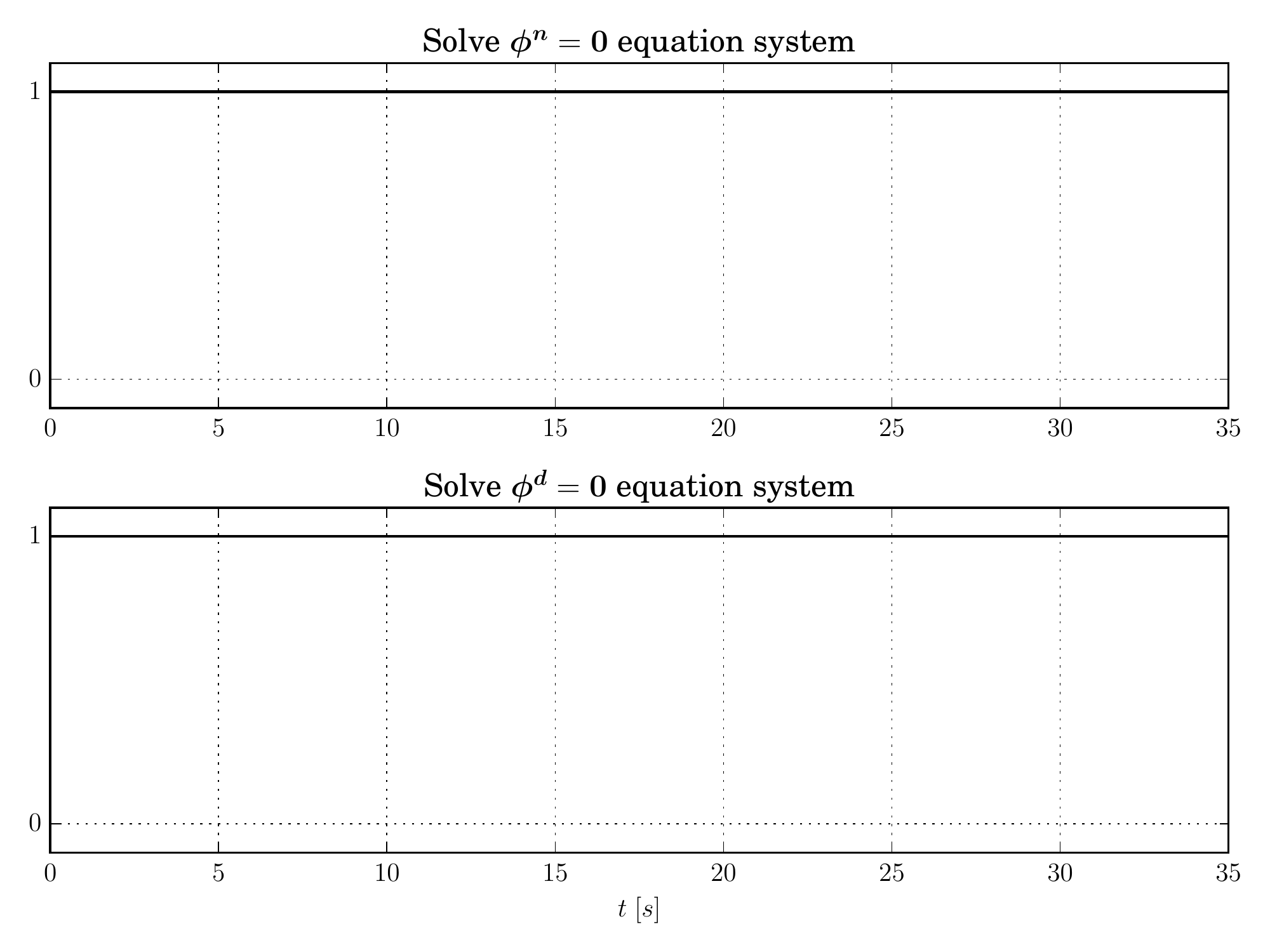}
  \caption{Iterative steps needed by the Newton-Rapshon algorithm}
  \label{fig:nriter}
\end{figure}

In Fig.~\ref{fig:nriter} the number of iterations required by the $\vq$-projection and $\vs$-projection steps are shown.
It is noticeable that the $\vq$-projection only requires a single Newton-Raphson\footnote{the tolerance used is $10^{-6}$ amounting to a negligible error of $\approx 10^{-3}~\text{mm}$ for lengths} iteration.
The same is true for the $\vs$-projection. This has required to integrate $\vs$ after the integration step using an explicit Euler procedure $\vs_{t+\Delta t}=\vs_t + \vds_t ~\Delta t$ leading to an to a smaller  error ($O(\Delta t^2)$)
at the start of the Newton-Raphson iteration. Clearly, the increased number of iterations is coincident with
the curved stretches. This result justify the approach adopted in which,  the $\vs$-projection is performed before the $\vq$-projection.

In Table~\ref{table:functions} the number of operations required for the evaluation of the different functions used by the proposed dynamic formalism are presented.
\begin{table}[h]
\centering
\caption{Atoms and operation for the evaluation of the model functions}
\label{table:functions}
\begin{tabular}{lrr}
\hline
Function     & Atoms & Operations \\ \hline
$\mM_{\vq\vq}$   & 1795  & 10910      \\
$\vdelta_{\vq}$  & 3648  & 19015      \\
$\vphin$     & 425   & 1709       \\
$\vphid$     & 541   & 2150       \\
$\mdPhindq$  & 784   & 4116       \\
$\mdPhinds$  & 846   & 6668       \\
$\mdPhiddq$  & 964   & 6574       \\
$\mdPhidds$  & 1113  & 8437       \\
$\vbeta^n$   & 0     & 0          \\
$\vgamman$   & 3779  & 25870      \\ 
$\frac{\partial \vdelta^{K}_{\vq}}{\partial \mathbf{f_i}^{K}} \mathbf{C}_i^{K}\frac{\partial \boldsymbol{\nu}_i}{\partial \vdq}$   & 350   & 1917       \\ \hline
\end{tabular}
\end{table}

The results show a correlation between the time for function evaluations and the number of operations. 
As a major result of this study, it can be seen that using the symbolic procedures proposed the penalties incurred for
using an exact treatment of the linear Kalker contact model are barely noticeable. Note that this is a fair comparison,
as the operation count related to other dynamic and kinematic computations are very optimized, showing numbers compatible with state-of-the-art
recursive formulations. This
puts into perspective the relevance of symbolic methods proposed in achieving hard real-time performance in the railway dynamics simulation context.

Still, there are still some possibilities to further improve the results given in this article.

1.-The dynamic system structure shown in Fig.~\ref{fig2} shows a decent amount of sparsity. This sparsity is shared with the IMEX discretized dynamic matrix.
Important savings can therefore be obtained using a sparse $LU$ algorithm.

2.- In the $\vs$-projection and $\vds$-projection problems $\mdPhidds$ is a maximum rank block-diagonal matrix with
$4\times4$ blocks \cite{samin2003}.  Therefore, its computation can be speeded up by big integer factor. The solution could be easily
implemented symbolically or even  in parallel.

3.- Removal of the repeated evaluation of constant atoms from the symbolic functions and reuse of atoms common to different exported functions.

As commented in the introduction, at the expense of some accuracy, partial linearization \cite{Escalona2015} or
base parameter reduction \cite{Iriarte2015}, can be used to further improve the computational performance of the model.

\section{Conclusions}
\label{sec5}
The purpose of the article was to test state-of-the-art methods for the symbolic modeling
 in the railway context. A complex locomotive running on a track with a complex and general surface geometry has been modeled and tested.
 
 Main aspects of the symbolic methods proposed are summarized: atomization, recursive operators, points and bases structures, general parameterization, etc.
Based on this methods the model is obtained using a direct implementation of the principle of virtual work.
Creep forces and moments are modeled using a direct symbolic implementation of the linear Kalker  model  without simplifications.
An Implicit-Explicit (IMEX) integrator has been proposed to cope with the contact model while attaining real-time performance.
The resulting equations are solved using coordinate partitioning MSD procedures.

A very stable hard-real-time-compatible performance with a time step of $1~\text{ms}$  is obtained. A CPU time of $256~\mu\text{s}$ per time step 
is required in a seven year old \textit{Intel Core vPro i5 @ 3500} MHz. 
It is noticeable the small time required for the determination of the creep forces
when an exact implementation of the linear Kalker model is used. Also, the compromise efficiency/robustness of the IMEX integrator proposed is remarkable.

The results obtained show the relevance of the
 methods proposed for the real-time simulation of railway vehicles.

There are still obvious possibilities to improve on the results presented in this work: better sharing of atoms,
constant atom revaluation, sparse linear solver implementation and parallelization, are the most obvious.
On top of this, with a small accuracy
penalty, techniques such as partial-linearization and parameter reduction can be used
to improve even further the results presented.

\section{Acnowledgements}
This work was partially supported by the ``Plan Nacional de Investigación Científica, Desarrollo e Innovación Tecnológica'', ``Ministerio de Economía y Competitividad'' [grant numbers IPT-2011-1149-370000 and TRA2014\_57609\_R].

\bibliographystyle{unsrt} 
\bibliography{bibliografia} 
\end{document}